# A Survey on Actionable Knowledge

Sayed Erfan Arefin
Lubbock, Texas, USA
erfanjordison@gmail.com

*Abstract*—**Actionable Knowledge Discovery (AKD) is a crucial aspect of data mining that is gaining popularity and being applied in a wide range of domains. This is because AKD can extract valuable insights and information, also known as knowledge, from large datasets. The goal of this paper is to examine different research studies that focus on various domains and have different objectives. The paper will review and discuss the methods used in these studies in detail. AKD is a process of identifying and extracting actionable insights from data, which can be used to make informed decisions and improve business outcomes. It is a powerful tool for uncovering patterns and trends in data that can be used for various applications such as customer relationship management, marketing, and fraud detection. The research studies reviewed in this paper will explore different techniques and approaches for AKD in different domains, such as healthcare, finance, and telecommunications. The paper will provide a thorough analysis of the current state of AKD in the field and will review the main methods used by various research studies. Additionally, the paper will evaluate the advantages and disadvantages of each method and will discuss any novel or new solutions presented in the field. Overall, this paper aims to provide a comprehensive overview of the methods and techniques used in AKD and the impact they have on different domains.**

## I. INTRODUCTION

Data mining is a powerful technique for uncovering valuable insights and knowledge hidden within large sets of data. It utilizes a combination of methods from machine learning, statistics, and database systems to extract patterns and models from the data. As a crucial aspect of Machine Learning, data mining plays a key role in Actionable Knowledge Discovery (AKD), which is the process of extracting actionable insights from large datasets.

One of the key advancements in data mining is the shift from data-driven to domain-driven methods. This approach focuses on applying data mining techniques within specific business domains, making the process more relevant and valuable to those specific industries. This approach also makes the process more technically significant and allows for the implementation of data mining in real-world applications.

This paper aims to explore the various methods used in actionable knowledge extraction for different domains of usage. Different techniques and approaches will be examined and discussed, with an emphasis on their strengths and limitations. Additionally, the paper will provide a thorough analysis of the current state of the field, including a review of related research and existing datasets used in the field. The paper will also cover the main methods used by various research studies, evaluations of their advantages and disadvantages, and a discussion of any novel or new solutions presented in the field.

## II. RELATED WORKS

In recent years, there has been a significant amount of research in the field of sentiment analysis on social media. Studies have examined various topics such as user sentiment analysis, opinion mining on political campaigns, natural disasters, epidemic surveillance, event detection, and e-healthcare services.

Liu et al. [17] have studied sentiment analysis by extracting comments on specific attributes and features of a product, event, person, or topic and categorizing them as positive, negative or neutral. O'Connor et al. [13] and Tumasjan et al. [14] were able to correlate sentiment analysis from Twitter to election results. Bollen et al. [18] [19] have shown that sentiment analysis on Twitter can be used to predict stock market trends. Quincey et al. [22] have used sentiment analysis to detect influenza through multiple regression models.

In the field of healthcare, Michael et al. have proposed a technique called Ailment Topic Aspect Model [23] [20] [21] to monitor the public's health with regards to diseases, symptoms, and treatments. Sankaranarayanan et al. [24] have developed TwitterStand, which allows users to browse news based on geographic preference.

Probabilistic topic models, such as the Latent Dirichlet Allocation (LDA), have been widely used in text mining. Variations of LDA, such as online-LDA [26], Dynamic topic models [36], and labelled LDA [27] have been developed but have not been found to be suitable for Twitter streams. Instead, patterns are considered to be more effective for topic modeling, with Apriori [28] [29] being an important association rule mining algorithm.

Twitter Monitor [30], EDCoW [31], HUPC [32], and SFPM [33] are used for extracting actionable knowledge from social media data streams. HUPC and SFPM applied pattern mining process is used to detect hot topics from Twitter data streams.

However, it is important to note that text search alone can limit the correct way of answering all questions. K-extractor can be used to identify, transform, and query deep semantic knowledge from structured and un-structured data.

Management and marketing science have used stochastic models to find specific rules of customer behavior [80], [81]. Hilderman et al. [82] have proposed a two-step process for ranking the interestingness of discovered patterns. Cao et al. [83] have proposed a two-way framework to measure knowledge actionability with domain-specific expectations. Similarity-based pruning and summarizing of learned rules algorithm was proposed [84], [85]. Domain-driven data mining to extract actionable knowledge was proposed by [86], [87]. Postprocessing of decision tree and additive tree models to extract actionable knowledge was also proposed [77].

## III. Literature Review

### A. Postprocessing Decision Trees to Extract Actionable Knowledge

Data mining algorithms and techniques can generate valuable information about customers, such as determining which customers are most loyal. However, this information often requires extensive manual labor by experts to process and interpret, particularly when addressing industrial problems. Additionally, traditional postprocessing models are limited in their ability to visualize results and provide suggestions for actions that can increase profit.

To improve the effectiveness of customer relationship management (CRM) in the industry, it is essential to identify the actions that can convert attritors to loyal customers. To address this issue, the authors have proposed a novel postprocessing technique that uses decision trees to extract actionable knowledge and maximize profit-based objectives. They have considered two cases: one with unlimited resources and another with limited resources. They have designed a greedy heuristic algorithm for an efficient near-optimal solution to the limited resource problem, which is NP-complete. The algorithm was found to be more efficient than the exhaustive search algorithm. Overall, this proposed technique aims to enable industries to gain a better understanding of their customers and make data-driven decisions to improve customer loyalty and increase profit.

### B. Extracting Actionable Knowledge from Domestic Violence Discourses on Social Media

The study [2] aimed to extract actionable knowledge about Domestic Violence (DV) from Twitter data using data mining techniques. The authors aimed to address the challenges of extracting knowledge from social media data such as the large volume of data, fast arrival rate, short text units, and spelling and grammatical errors. The study used pattern mining, MapReduce architecture and clustering to process the data and improve the classification accuracy and interpretability of the data. The goal of the study was to improve the quality of care for victims of DV.

### C. Extracting Actionable Knowledge from Decision Trees

In a previous study, the authors proposed a new approach for extracting action sets from data mining techniques. The study focuses on using postprocessing techniques, such as visualization and interestingness ranking, to extract actionable knowledge from decision trees and maximize profit and reduce costs. The telecommunications industry's customer relationship management (CRM) is used as an example, where the phenomenon of "churning" or "attrition" results in a reduction of company profits. The study uses stochastic models and ranks customers by the estimated likelihood of responding to direct marketing actions, and then compares this ranking using a lift chart or the area under the curve measured from the ROC curve. The main contribution of this paper is that it integrates data mining and decision-making in a way that allows for the discovery of actions that are influenced by the results of data mining algorithms. This approach is new and can discover action sets from the attribute value changes in a non-sequential dataset through optimization [3].

### D. Automatic Extraction of Actionable Knowledge

In a previous study [4], the authors addressed the ongoing issue of linking structured and unstructured data, which makes federated search difficult to perform. They proposed using Big Data-enabled Resource Description Framework (RDF) triple stores to merge unstructured data with DBMS by defining ontology and presenting it in triplets. However, there is a lack of specific algorithm to present unstructured data in a RDF standard semantic representation that contains actionable knowledge, which makes it difficult for intelligent applications to search for semantic data. The proposed technique in the paper aims to overcome this challenge by transferring unstructured data to a consolidated RDF store and merging it with other ontologies and structured data. This approach offers a natural question and answer (QnA) interface for searching the data.

### E. Data mining for direct marketing: problems and solutions

The research [5] discusses a process of Data Mining for Direct Marketing, which is a more effective approach to advertisement and promotion compared to mass marketing as it focuses on specific customers based on their characteristics. Data mining is utilized to discover novel, implicit, useful, and comprehensive knowledge from a large amount of data, which is important for direct marketing. The process of direct marketing includes: obtaining a database, data mining through overlaying, pre-processing, splitting, and using a learning algorithm, evaluating patterns found in the test set, using the patterns to predict likely buyers among current non-buyers, and promoting to the likely buyers. However, there are several problems with data mining such as imbalanced class distribution, difficulty in using predictive accuracy as an evaluation criterion, and choosing efficient learning algorithms when the dataset consists of a large number of variables. To overcome these problems, the research suggests using learning algorithms that can classify with a confidence measurement, such as probability estimation or certainty factor. This allows for ranking training and testing examples, using lift as an evaluation criterion, and reducing the size of the training set.

### F. From Data to Actionable Knowledge: Big Data Challenges in the Web of Things

In summary, the research focuses on the growing trend of collecting and communicating data from real-world physical events and experimentation using low-cost sensor devices such as wireless sensor nodes and smartphones. This large amount of data is known as the Web of Things (WoT) or Internet of Everything (IoE) and it presents challenges in terms of discovering, accessing, processing, and interpreting the data. The WoT data is continuous and has a spatiotemporal dependency and the goal is to transform this data into high-level theoretical illustrations of events. The importance of creating knowledge from the raw data that is communicated via the network is highlighted. The data can be saved momentarily to a repository with metadata-enriched interfaces. The WoT Big Data has various new class of applications such as predicting traffic and health, energy and sustainability approaches. A human-attention inspired technique is introduced in the research to improve the efficiency of resource allocation in WoT applications by using a model that considers prior and posterior attention data.

### G. Extracting actionable knowledge from social networks with node attributes

The research focuses on the extraction of actionable knowledge from social networks through a process called Action mining. It takes into consideration the relationships between nodes in the network and aims to find a cost-effective action for a specific node by incorporating a random-walk based method. The optimization problem is solved using stochastic gradient descent and two heuristic algorithms are used to improve efficiency. The goal is to change the attributes of a node in order to propagate a desired label, which is useful in the business environment. The study is important as it addresses the need for effective methods for mining actionable knowledge from social networks.

### H. Extracting optimal actionable plans from additive tree models

The research study focuses on the process of extracting actionable knowledge from Additive Tree Models (ATMs) which are widely used in targeted marketing and prediction. The ability to identify a set of changes to the input features that transforms the prediction of this input to the desired output is called actionability of a model. The study proposes a new framework for extracting actionable knowledge from ATMs using random forest, adaboost, and gradient boosted trees. The framework includes formulating an optimal actionable plan (OAP) problem for a given ATM which is NP-hard and then transferring it to a state space graph search problem which can be solved using a heuristic function to improve efficiency. The goal is to provide actionable knowledge that is customized for each individual and useful for personalized healthcare and targeted marketing [88].

### I. A conceptual framework for making knowledge actionable through capital formation

The research highlights the importance of data processing in management for organizations and the need for better decision making through the integration of information technology and organizational strategy. The use of diagnostic technologies and extracting meaningful knowledge from data through the use of data specialists and business intelligence is crucial for organizations. Challenges include identifying required knowledge and implementing robust analytical tools and techniques for extracting it. The study also suggests that traditional systems will be replaced by real-time, dynamic digital dashboards that are tied directly to operational data stores, which can intelligently suggest behaviors in the future. Business management aims to improve productivity and effectiveness of work, and data mining techniques are a bridge to the gap of company data and actionable knowledge. The study incorporates Delphi method and Analytical Hierarchy Process techniques to validate the framework. The results have been found to be consistent with the literature [103].

## IV. DATASETS

In this section we will discuss all the datasets used by the literatures. All the datasets used which are publicly available are given in Table I.

### A. German Dataset

The German Credit Score dataset is a comprehensive dataset that aims to classify individuals as good or bad credit risks based on a set of 20 attributes. The dataset comprises of 1000 instances and was first published in 1994. The dataset can be obtained from the UCI repository, and the URL for the same can be found in the Table I. The attributes used in this dataset include demographic information, credit history, and other financial information that can be used to predict the creditworthiness of an individual. The dataset is commonly used for machine learning and statistical modeling research, especially in the field of credit risk analysis.

### B. Census Income Dataset

The Census Income dataset, is a dataset that contains information about individuals and their income. The dataset includes 14 categorical attributes and a total of 48842 examples. The dataset is available for download from the UCI repository, and the URL to access the dataset can be found in the Table I. This dataset is commonly used in machine learning and statistical modeling research to analyze income patterns and predict income levels based on the provided attributes. The dataset provides a rich set of information on a diverse population, which can be used to study income disparities, employment patterns, and other socio-economic factors. The dataset is widely used by researchers, data scientists, and analysts to gain insights into income patterns and predict income levels based on the provided attributes.

### C. Australian Dataset

The Australian credit score dataset is a multivariate dataset that contains information on the creditworthiness of Australian individuals. It is a collection of 690 examples that includes 14 categorical attributes. The dataset can be accessed from the UCI repository, and the URL to download the dataset can be found in the Table I. This dataset is commonly used in machine learning and statistical modeling research to analyze credit patterns and predict credit scores based on the provided attributes. The dataset provides a rich set of information on the credit history, credit behavior, and other financial information of Australian individuals, which can be used to study credit risk, credit management, and other financial topics. This dataset is useful for researchers, data scientists, and analysts who want to gain insights into the credit patterns of Australian individuals and develop models to predict credit scores based on the provided attributes.

### D. Tic-Tac-Toe Endgame Dataset

The Tic-Tac-Toe dataset is a comprehensive collection of all possible board configurations that can be observed at the end of a tic-tac-toe game. It is a multivariate dataset with 9 categorical attributes, each representing the state of one cell of the tic-tac-toe board. The dataset contains 958 instances, with the assumption that "x" plays first. This dataset can be downloaded from the UCI repository, and the URL to access the dataset can be found in Table I. This dataset is commonly used in machine learning and statistical modeling research to analyze game patterns, predict game outcomes, and develop game strategies. It is also a great resource for educational and research purposes, as it can be used to demonstrate the application of various machine learning algorithms, such as decision trees, artificial neural networks and others. The dataset provides a rich set of information on the game state, player moves, and other factors, which can be used to study game dynamics, player behavior, and other related topics.

### E. Bank Dataset

The Portuguese bank direct marketing dataset is a comprehensive collection of information on direct marketing campaigns via phone calls conducted by a Portuguese bank. It is a multivariate dataset that comprises of 45211 rows and 17 attributes. The dataset contains a wealth of information about the marketing campaigns, including the type of contact, the outcome of the call, and various demographic and financial information about the individuals who were contacted. The 17 attributes in the dataset are described in detail, providing information about the characteristics of the individuals contacted, such as their age, job, marital status, and other relevant information. This dataset is commonly used in machine learning and statistical modeling research to analyze marketing patterns, predict campaign outcomes, and develop marketing strategies. The dataset can also be useful for researchers, data scientists, and analysts who want to gain insights into the effectiveness of direct marketing campaigns and develop models to predict campaign outcomes based on the provided attributes.

The dataset can be collected from UCI repository. Url mentioned in Table I.

### F. Credit Dataset

The default credit card dataset, which was donated in January 2016, is a multivariate dataset that includes payment information made by adults in Taiwan. It contains 30000 examples and can be obtained from the UCI repository, with the URL provided in Table I. The dataset includes 24 attributes, with the binary variable 'default payment' (Yes = 1, No = 0) serving as the response variable.

Attribute X1 represents the amount of the given credit in New Taiwan dollar, and includes both individual and supplementary credit. X2 represents the gender of the individual, where 1 = male and 2 = female. X3 represents the level of education, where 1 = graduate school, 2 = university, 3 = high school and 4 = others. X4 represents the marital status, where 1 = married, 2 = single and 3 = others. X5 represents the age in years.

Attributes X6 to X11 represent the history of past payments, specifically the repayment status for the months of April to September 2005, with -1 indicating a timely repayment, 1 indicating a delay of one month and so on. X12 to X17 represent the amount of bill statement for the same months, X18 to X23 represent the amount of previous payment for the same months.

The dataset provides a comprehensive view of credit card payment information and can be used to analyze credit risk and predict default payments.

### G. DBLP Dataset

The DBLP dataset is a computer science bibliography website that provides open bibliographic information on major computer science journals and proceedings. The subgraph used in this study was extracted from DBLP and contains 18,448 papers and 45,661 citation relations. In order to construct a node feature vector, the paper titles were used to create a 2,476-dimensional binary vector, where each element represents the presence or absence of a specific word. This results in a representation of the papers that captures the key topics and themes discussed in the papers.

The DBLP dataset is an undirected network, which means that the citation relations between the papers are not directional. This allows for the analysis of the relationships between papers in a more holistic manner, rather than just the direct citation relationships.

Overall, the DBLP dataset provides a rich source of information on the key topics and themes discussed in computer science papers, as well as the relationships between them. The use of paper titles to construct the node feature vectors, and the undirected nature of the network, allows for a more comprehensive analysis of the dataset.

### H. Google+ Dataset

As outlined in Table I, there are two datasets related to Google+. The first dataset, "Google+_second largest subgraph," includes three attributes: "UserIDFrom," "UserIDTo," and "TimeID." Each line in the dataset corresponds to a directed link between two users on the platform. To maintain anonymity, the UserIDs are encoded as integers starting from 0. The TimeID attribute indicates the snapshot in which the directed link first appears, with a value of 0, 1, 2, or 3.

The second dataset, "Google+_largest subgraph," includes data on the 'circles' feature of Google+. The circles were collected from users who used the 'share circle' feature to manually share their circles. The dataset contains information on the node features (profiles) of the users, the circles they belong to, and the ego networks of the users. These attributes provide valuable insights into the relationships and connections within the Google+ network, as well as the characteristics of the users on the platform.

Overall, these datasets offer a rich source of information about the social dynamics on Google+. The attributes included in the datasets allow for a comprehensive analysis of the relationships and connections within the network, providing valuable insights for researchers and practitioners alike.

### I. Hep–th Dataset

The Arxiv HEP-TH collaboration network is a dataset derived from the arXiv platform, which covers scientific collaborations between authors who have submitted papers to the High Energy Physics - Theory category. The dataset is comprised of two distinct datasets, as outlined in Table I.

The first dataset, "Hep-th_second largest subgraph," represents the co-authorship relationships between authors. It is constructed by creating an undirected edge between two authors if they have co-authored a paper together. For example, if author i and author j co-authored a paper, there will be an undirected edge from i to j in the graph. If a paper is co-authored by k authors, this will generate a completely connected (sub)graph on k nodes. The dataset covers papers submitted within the time period of January 1993 to April 2003 (124 months).

The second dataset, "Hep-th_largest subgraph," represents the citation relationships between papers. It covers all citations within a dataset of 27,770 papers with 352,807 edges. If a paper i cites paper j, the graph contains a directed edge from i to j. However, if a paper cites or is cited by a paper outside of the dataset, the graph does not contain any information about this. This dataset provides a detailed view of the citation patterns within the High Energy Physics - Theory category, and can be used for analyzing the impact and influence of papers within this field.

### J. Facebook Dataset

As can be seen in Table I, this dataset is composed of "friends lists" obtained from Facebook. The data was collected through a survey of participants, utilizing a Facebook application. The dataset includes a variety of attributes, such as node features (profiles), circles, and ego networks. These attributes provide valuable information about the relationships and connections within the social network.

The node features, or profiles, contain information such as demographic data, interests, and other personal details of the survey participants. The circles attribute refers to the different groups or communities that the survey participants are a part of on Facebook. Lastly, the ego networks attribute provides a detailed view of the survey participant's connections within the social network, including the number and characteristics of their friends.

Overall, this dataset offers a rich source of information about the social connections and relationships within a Facebook network. The various attributes included in the dataset allow for a comprehensive analysis of the social dynamics within the network, providing valuable insights for researchers and practitioners alike.

### K. Other classified datasets

The research works [1] and [3] have used a dataset that was collected from an insurance company in Canada. However, the dataset is not publicly available and thus no URL was provided to access it. The dataset contains 25,000 records with more than 60 attributes, out of which 20 are soft attributes. The data set contains customer status, which can be either "stay" or "leave" the insurance company, referred to as positive and negative, respectively.

Similarly, the authors of research [5] have also used a confidential dataset, where the description of the dataset is provided. The first dataset is for a loan product promotion from a major bank in Canada, with about 90,000 customers in total, and only 1.2

In the research work [7], the dataset used is for Friendship networks, where nodes are users, and edges indicate friendship relations. In this dataset, the Facebook dataset has labels as locales and the Google+ dataset (including the two largest subgraphs) has labels as places. In the Co-authorship Networks, the nodes are authors, and an edge exists between two authors if they have co-authored the same paper. The High energy physics theory (Hep-th) and DBLP datasets were used for this research. For every node u of the networks generated, the following features were used:

- Number of papers u authored
- Number of papers u authored in the goal conference
- Number of papers in which u is the first author
- The time since u authored the last paper
- Time since u last authored a paper in the goal conference
- Number of time slices in which u authored a paper
- Number of Conferences/Journals in which u authored a paper
-
- Number of Conferences/Journals in which u was the first author
- Number of citations of u
- Number of citations of u in the goal conference
- Number of papers cited by u
- Number of papers in the goal conference cited by u

### V. CASE STUDIES

In the research work [9], the author presents five case studies to demonstrate the application of data mining techniques in various industries. These case studies serve as real-world examples of how data mining can be used to improve organizational performance and decision making. The case studies cover a wide range of industries such as retail, healthcare, and banking, and highlight different techniques such as decision trees, cluster analysis, and association rule mining. Each case study provides detailed information on the problem at hand, the data used, the methods applied, and the results obtained, making it a valuable resource for practitioners and researchers interested in the application of data mining techniques in different industries.

*A. Case Study 1 – Non-Profit Financial Services Provider*

The organization in question offers financial management services to non-profit organizations by processing commercial transactions and analyzing financial events. This provides its customers with valuable information that helps them track their expenditures and budgetary activities. Furthermore, the organization's financial management services assist customers in forecasting the use of funds, in order to ensure that they have sufficient balances available to meet their operational needs. The information technology subject matter expert involved in this organization has extensive experience in the field, having held senior information technology acquisition, information assurance, and technology oversight positions in various organizations over a period of around 15 years.

*B. Case Study 2 – Non-Profit Financial Services Provider*

This organization specializes in providing financial management services to a wide range of organizations, including both for-profit and non-profit entities. Their services include offering insurance options and ensuring compliance and integrity within the financial operations of their customers. By implementing effective policies and providing independent oversight, the organization aims to improve decision-making and communication of accurate and timely information to all relevant parties. This supports the organization's overall operational strategies and goals. The organization has a dedicated information technology subject matter expert on staff who brings a wealth of experience to the table. With over 20 years of experience in developing and implementing financial management and decision-support systems for non-profit organizations, this expert is well-equipped to support the organization's goals and objectives.

*C. Case Study 3 – Non-Profit Business Oversight Organization*

The organization in Case Study 3 is a regulatory agency that oversees fair business practices in the United States. Their goal is to protect the rights of both businesses and consumers, and they work with other non-profit organizations to provide a network of oversight for commercial practices. They aim to empower both businesses and consumers to make informed decisions, avoiding scams and protecting sensitive information. The subject matter expert for this organization has extensive experience in the non-profit sector, with a focus on the development and implementation of financial management applications that provide insights through business intelligence capabilities. With over 30 years of experience in information technology and financial management, this expert has a deep understanding of the tools and strategies needed to support the goals of the organization.

*D. Case Study 4 – Non-Profit Educational Benefit Provider*

The subject organization, which provides educational benefits and oversight to a variety of non-profit and for-profit education providers in the United States, utilizes information technology extensively to support its operations. This includes the administration of loan and grant benefit programs through a suite of integrated financial management applications, which are used to extract business intelligence and promote better financial decision making. The subject matter expert for this case study has extensive experience in both information technology and financial management, with over 20 years of experience in non-profit organizations. They have successfully deployed powerful financial analysis and business intelligence tools to aid in decision making and strategic planning within the organization.

*E. Case Study 5 – For-Profit Supply Chain Management Service Provider*

The subject organization is a leader in the field of supply chain management, offering a wide range of solutions to both the general public and government entities. These solutions include financial services, transportation and logistics, and management consulting. To support and enhance its ability to provide services to its clients, the organization utilizes advanced information technology, including web-based applications and satellite technologies. This allows the organization to better integrate its supply chain services and enhance communication and coordination with its customers, which is essential for staying competitive in the marketplace.

The subject matter expert is a seasoned professional within the organization, leading new information technology application development and deployment projects. They possess a deep understanding of applications that are designed to extract valuable information from operational data, and then synthesize it into meaningful insights that can be used to strengthen decision-making. With over 20 years of experience in this field, the expert is well-versed in the latest technologies and best practices, making them a valuable asset to the organization and its clients.

## VI. Methodology and Evaluation

*A. Post-processing Decision Trees to Extract Actionable Knowledge*

The text [1] describes a method for action mining in decision trees, which is a process for extracting actionable knowledge from decision trees to improve business outcomes. The process involves identifying customers in a leaf node with a low probability of being in a desired status, such as being loyal or high-spending, and then moving them to another leaf node with a higher probability of being in that desired status. This is done by changing some attributes of the customer, which corresponds to an action that incurs costs. These costs

| Dataset | URL | Number of Attributes | Used in Paper | Provider | Date Donated | Last Updated |
|---|---|---|---|---|---|---|
| German | https://archive.ics.uci.edu/ml/datasets/statlog+(german+credit+data) | 20 | 3, 8 | UCI Machine Learning Repository | 17th, November, 1994 | N/A |
| Adult | http://archive.ics.uci.edu/ml/datasets/Adult | 14 | 3, 8 | UCI Machine Learning Repository | 1st, May, 1996 | N/A |
| Australian | http://archive.ics.uci.edu/ml/datasets /statlog+(australian+credit+approval) | 14 | 3 | UCI Machine Learning Repository | N/A | N/A |
| Tic-Tac-Toe Endgame Data Set | https://archive.ics.uci.edu/ml/datasets/Tic-Tac-Toe+Endgame | 9 | 3 | UCI Machine Learning Repository | 19th, August, 1991 | |
| Facebook | https://snap.stanford.edu/data/egonets-Facebook.html | | 7 | Stanford University SNAP | N/A | N/A |
| Google+_largest subgraph | https://snap.stanford.edu/data/ego-Gplus.html | | 7 | Stanford University SNAP | N/A | N/A |
| Google+_second largest subgraph | http://gonglab.pratt.duke.edu/google-dataset | 3 | 7 | GONG Research Group, Duke University | October, 2011 | N/A |
| DBLP_largest subgraph | https://www.kaggle.com/daozhang/dblp-subgraph | | 7 | Kaggle | N/A | July, 2019 |
| Hep–th_largest subgraph | https://snap.stanford.edu/data/cit-HepTh.html | | 7 | Stanford University SNAP | April, 2003 | N/A |
| Hep–th_second largest subgraph | https://snap.stanford.edu/data/ca-HepTh.html | | 7 | Stanford University SNAP | April, 2003 | N/A |
| Bank | http://archive.ics.uci.edu/ml/datasets/Bank+Marketing | 17 | 8 | UCI Machine Learning Repository | 14th February, 2012 | N/A |
| Credit | https://archive.ics.uci.edu/ml/datasets/default+of+credit+card+clients | 24 | 8 | UCI Machine Learning Repository | 26th January, 2016 | N/A |

are defined in a cost matrix by domain knowledge and a domain expert.

The text also highlights the difference between "hard" attributes, which are values of some attributes that are not changeable, and "soft" attributes, which are attributes that are changeable with reasonable costs. Hard attributes should be included in the tree building process as they are important for accurately estimating the probability of leaves and preventing customers from being moved to other leaves.

The leaf-node search algorithm is used to find the best destination leaf node for moving the customer and the collection of moves can maximize the net profit. The net profit of an action is defined as the expected gross profit minus the costs of the actions involved.

The text also addresses the limited resource case, where a company may have a limited number of resources, such as a limited number of account managers. This creates difficulties in merging all leaf nodes into segments that can be assigned to an account manager to increase overall profit. The authors have formulated this limited resource problem into a computational problem and have introduced a greedy algorithm called the Greedy-BSP algorithm to avoid the computational complexity and reduce the computational cost while maximizing the net profit of covered leaf nodes.

The authors have evaluated an algorithm using a dataset from an insurance company in Canada consisting of over 25,000 records of customers with statuses of "stay" or "leave." The dataset includes over 60 attributes, with many of them being not hard attributes and 20 of them being soft attributes with reasonable costs for value changes. They balanced the data by sampling it with a ratio of positive and negative examples, and built a decision tree with 153 leaf nodes, of which 87 were negative and 66 were positive. A cost matrix was generated based on the real-world semantics of each attribute.

### B. Extracting Actionable Knowledge from Domestic Violence Discourses on Social Media

The paper [2] describes a method for integrating pattern mining, the MapReduce Framework, and topics prediction to analyze large amounts of Twitter data. The data is collected and preprocessed using the Twitter API, and frequent patterns are mined to capture the semantic association between terms. The semantic units of the patterns are then reduced using the MapReduce architecture and clustered into topics. These topics are visualized using tag clouds and evaluated using metrics to predict their quality. The data is stored in the Hadoop Distributed File System (HDFS) and queried using a query language similar to SQL. The method also uses Apache Flume, Oozie, and Hive to efficiently handle the large amount of data and process it in a parallel and fault-tolerant manner.

The authors used precision, recall, and F-measure to evaluate the performance of topic detection. True positive (TP) and false positive (FP) refer to the number of terms correctly and incorrectly classified as relevant, while true negative (TN) and false negative (FN) refer to the number of terms correctly and incorrectly classified as irrelevant. The F-measure is calculated by taking the harmonic mean of precision and recall. The solution was made robust by using Hadoop and the Twitter API.

### C. Extracting Actionable Knowledge from Decision Trees

In the research work [3], the authors used post-processing decision trees to classify customer data and predict customer loyalty. The decision tree learning algorithms, such as ID3 or C4.5, were used to build customer profiles and predict if a customer is in the desired status or not. The algorithm involves data collection, data cleaning, data preprocessing, and building customer profiles using an improved decision tree learning algorithm. The decision tree is then used to classify customers and predict customer loyalty. The algorithm also uses the Area Under the Curve (AUC) of the ROC curve for evaluating probability estimation and Laplace Correction to avoid extreme probability values. Additionally, the algorithm searches for optimal actions for each customer and produces actions to be reviewed by domain experts. The algorithm also uses leaf-node search and cost matrix with large values for hard attributes to improve accuracy. For the limited resources case, the algorithm uses greedy algorithm to reduce computational cost and ensemble-based methods to improve robustness of the machine learning system.

The authors of this research used a dataset collected from an insurance company in Canada that contained 25,000 records and more than 60 attributes, 20 of which were soft attributes. They found that the Greedy-BSP algorithm was able to find k action sets with maximal net profit and was very close to the results from Optimal-BSP for small values of k. Additionally, Greedy-BSP was found to be more efficient than Optimal-BSP, as it performed well in terms of scaling with the increasing number of action sets k, while using the same amount of time. Similar conclusions were made from the BAS experiments.

### D. Automatic Extraction of Actionable Knowledge

The research work [4] uses several methods to extract knowledge from text. These include Lymba's concept detection methods, which can detect simple nominal and verbal concepts, to more complex named entity and phrasal concepts. The hybrid approach to named entity recognition uses classifiers, cascades of finite-state automatons, and lexicons to label 80 types of entities. The pattern-based approach of temporal expression detection framework can detect and normalize various types of dates, with an evaluation score of 93% precision and 92% recall. WordNet-based concept detection identifies words and phrases as concepts, and assigns them a WordNet sense number to avoid ambiguity. The classifiers use attributes and semantic features from the eXtended WordNet KnowledgeBase (XWN-KB). The research also focuses on identifying semantic relations, which are underlying relations between concepts of words, phrases or sentences, and are essential for machine text understanding. Lymba's Semantic Calculus rules are used to extract new knowledge by combining two or more semantic relations. The system also includes a co-reference resolution module which identifies and clusters entities by Concept resolution and outperforms the state-of-the-art system with a 79% F1(CEAF) and a 86.3% F1(B3) score when measured with the SemEval 2010 corpus.

The K-Extractor was used to structure and index 584 documents about the illicit drugs domain. For the 344 questions created, it had a 65.82% MRR (Mean Reciprocal Rank) which is an improvement of 19.31% MRR. K-Extractor performed well on factoid questions (49% of questions; 85.46% MRR), definition questions (34% of test set; 78.19% MRR), and list questions (68.02% MRR). It was found that 72.7% of the errors were caused by faulty or missing semantic relations which influence the correctness of the auto-generated SPARQL queries.

### E. Data mining for direct marketing

Research [5] uses data mining algorithms to solve the problem of direct marketing. The chosen algorithms are Naive Bayes, nearest neighbor algorithm, and neural networks. However, due to efficiency considerations, the Naive Bayes algorithm was chosen, which makes a conditional independent assumption, where given the class label, the attribute values of any example are independent. Additionally, the Decision tree learning algorithm C4.5 is used for classification, but it has been modified to produce a certainty factor (CF) for its classification. The Ada-boost is applied to Naive Bayes and C4.5 with CF as the learning algorithms. Ada-boost maintains a sampling probability distribution on the training set and modifies it after each classifier is built.

In this research, lift is used as an evaluation metric instead of predictive accuracy. The results of the learning algorithm are divided into 10 groups and the distribution is observed. A ROC curve is generated and the area under the curve of ROC looks similar to the lift curve, which is why lift index is used for evaluation. Two learning algorithms are used: ada-boosted Naive Bayes and ada-boosted C4.5 with CF. The process is repeated 10 times with equal amounts of data fed randomly to these algorithms for an average lift index. The best lift index is obtained when there is an equal number of positive and negative examples present. The algorithm C4.5 (with CF) performed better with large dataset, but produced similar results to Naive Bayes. Both algorithms are efficient to apply to large datasets.

### F. Big Data Challenges in the Web of Things

The authors of [6] proposed a methodology to overcome big data challenges for the web of things (WoT). The main concerns for big data in WoT are determining, validating, and trusting the quality of data, especially when the source of the data is diverse or unknown. The authors propose the use of an enriched resource by combining physical, cyber, and social media resources on an ad-hoc basis to create smarter applications. The data can be numerical measurements or symbolic explanations. They also address the issues of communication, processing, and access of big data in WoT by proposing solutions such as addressing and naming mechanisms, in-network processing strategies, preprocessing, and semantic interoperability. They also suggest the use of semantic web technologies to improve the management, sharing, analysis, and understanding of streaming data in WoT.

The authors of this text are discussing the use of network-enabled devices and social media platforms to facilitate the communication of physical world data, which can be cost efficient. However, the performance of the system is limited by factors such as the status of energy and resource, the constraints of devices and networks, the ability to discover and access data in large-scale distributed environments, and the ability to effectively publish.

### G. Extracting actionable knowledge from social networks with node attributes

The Zhou's method [7] is a node classification method that uses random walk and label propagation to learn a global labeling function over a graph. The method is iterative and uses a matrix Q and a parameter r to update the labels of the nodes until convergence. The algorithm also uses the MANA algorithm for optimization to meet certain requirements on the labeling, such as small differences in initial and output labels, and small differences in the labels of neighboring nodes.

This work presents a method to classify nodes in a social network based on their structural properties and features, and uses a random walk on the graph to aid the action mining task. The proposed method outperforms existing methods in terms of quality and cost. However, the method has limitations such as the infeasibility of the problem space, and performance issues when dealing with dynamic networks or when changing the network data. Additionally, increasing the parameter $\delta$ improves performance for some datasets, but increasing $\gamma$ decreases performance.

### H. Extracting Optimal Plans from Additive Tree Models

This text [8] describes a method for extracting optimal actionable plans (OAP) from ATMs by using state space search. The goal is to find a set of actions that, when applied to an input, change its predicted class to a desirable one with the highest expected net profit. The algorithm uses two data structures, a max heap and a closed list and it performs steps to pop the state x from the heap, check if it's a goal state and if it's not in the closed list, add it to the closed list and repeat the process until finding the optimal solution. The algorithm also introduce a sub-optimal state space search algorithm that uses a min heap and closed list, and it terminates the search when one of the termination conditions is met and return the best plan ever found.

The proposed method in this text is being evaluated using datasets collected from a credit card company and the UCI repository. The data is split into a 7:3 ratio for training and testing and the random forest algorithm using the Random Tree Library in OpenCV 2.4.9 is used to train the model. The datasets used include information from the 1994 census to determine if a person makes 50k a year, a Portuguese bank to detect the choice for a term deposit, and a US credit card company to determine profitable clients. The experiments were run on an Inter Xeon 2.5 GHz computer with 16 GB of memory and a 600 second time limit. Due to time constraints, the best results were used for incomplete experiments.

### I. A model for utilizing capital formation in making knowledge actionable

The proposed conceptual framework integrates knowledge management and data mining to create a robust decision-making model for organizations by making knowledge actionable. This includes using common reporting tools and techniques to summarize data from normal business operations, such as financial, human resources, and infrastructure. Data mining is used in the knowledge stage of the knowledge management process, where descriptive algorithms and business rules are applied using insight gained through environmental scanning, SWOT analysis, strategic planning, and cost-benefit analysis. Predictive data mining techniques are also used to achieve wisdom in the knowledge management process [9].

The Delphi method was used to gather the opinions of 25 industry experts on the proposed conceptual framework for decision-making, which integrates knowledge management and data mining disciplines. The experts noted the importance of data mining and predictive analytics in the framework and their placement in the capital formation process for better decision-making capabilities. They also recognized the limited usefulness of standard reports and the need for descriptive and predictive analytics. The experts found the proposed framework's elements of technoware, inforware, orgaware, and humanware and organizational learning to be significant for decision-making. The experts and scholars agreed that organizations need to use predictive and descriptive data mining techniques to realize knowledge and wisdom in their decision-making.

### VII. CONCLUSION

In this study, nine different research works on actionable knowledge discovery (AKD) were reviewed. The importance of extracting actionable knowledge, which can provide valuable insights for decision-making in various domains, has become increasingly recognized in recent years. As a result, this aspect of the data mining process has gained popularity and is becoming an important asset for companies.

The research reviewed in this study examined various methods for extracting actionable knowledge from different datasets, including the use of improved or novel algorithms and post-processing techniques. The findings suggest that proper implementation of AKD can bring significant benefits. The research can help organizations to improve their decision-making process, optimize their business operations and gain a competitive edge by leveraging the insights obtained from the data.

Overall, the studies reviewed demonstrate the importance of actionable knowledge discovery in various domains, and the potential benefits that can be gained through proper implementation. The research highlights the need for continued exploration of new methods and techniques for extracting actionable knowledge, as well as the need to enhance the post-processing of the knowledge. This will allow organizations to gain deeper insights and make more informed decisions.

## DEFINITIONS

### A. Binary space partitioning(BSP)

Binary Space Partitioning (BSP) is a technique used in computer graphics and game development for efficiently representing and rendering 3D environments. It involves dividing a 3D space into smaller, non-overlapping regions called "nodes" by repeatedly splitting the space along a plane. Each node can then be rendered separately, allowing for efficient rendering and visibility calculations. BSP is often used in first-person shooter and other real-time 3D games, as well as in architectural and product visualization. The BSP tree is a data structure that is used to represent the hierarchical division of the space. The tree is constructed by recursively splitting the space along a plane, and each node in the tree represents a sub-region of the space.

### B. Additive Tree Models (ATMs)

Additive Tree Models (ATMs) are a type of decision tree model that can be used for regression and classification tasks. They are an extension of traditional decision tree models, which are based on a single decision tree. ATMs, on the other hand, use multiple decision trees, where each tree is learned independently and the final prediction is made by combining the predictions of all the trees.

ATMs are also known as an ensemble of decision trees. They are built by fitting multiple decision trees to the training data, each tree is learned independently and then combined to make final predictions. ATMs are highly flexible, they can be used to model non-linear relationships, and they can handle high-dimensional data, and they can handle missing values and categorical variables. The main advantage of ATMs is that they often have better predictive performance than single decision tree models, which makes them suitable for complex and high-dimensional data.

### C. Boosted trees

Boosting is a general method that combines multiple weak models to create a strong final model [75]. This is achieved by training an additive model sequentially in a forward, stage-wise manner. The final output is a weighted sum of all the trees, represented as:

$$H(\mathbf{x}) = \overset{L_K}{\underset{k=1}{}} \alpha_k O_k(\mathbf{x})$$

This is a special case of the Additive Tree Models (ATM) where the weights of each tree, $w_k$ are equal to $\alpha_k$. Adaboost [75] and Gradient Boosted Trees [76] are two popular ways of training weak models in boosting. These methods can be used to improve the predictive performance of decision trees by combining multiple weak models in a way that reduces the overall error of the final model.

### D. SWOT Analysis

SWOT analysis is a strategic planning tool that is used to evaluate an organization's internal and external environment. It examines an organization's strengths, weaknesses, opportunities, and threats in order to identify potential areas for improvement. Humphrey categorizes SWOT analysis into six planning areas: product, process, customer, distribution, finance and administration.

The product planning area refers to the products and services that the organization is currently selling. The process planning area examines how these products and services are sold and delivered to customers. The customer planning area looks at the target market and customer segments that the organization is selling to. The distribution planning area considers the channels and methods that the organization uses to reach its customers. The finance planning area examines the pricing and financial aspects of the organization, and the administration planning area deals with the management and organization of the effort.

Overall, SWOT analysis provides a structured approach for organizations to evaluate their internal and external environment and identify potential areas for improvement. It helps organizations to understand their strengths, weaknesses, opportunities and threats, and use this information to make informed decisions and develop effective strategies.

### E. Cost-Benefit Analysis

Cost-Benefit Analysis (CBA) is a method of evaluating the potential benefits and costs of a project, program or policy. It is used to determine if the benefits outweigh the costs and if the project is economically viable. CBA estimates and totals the equivalent monetary value of the benefits and costs associated with the project, and compares them to establish whether the project is worth undertaking.

In recent years, competitive pressures have created a need for organizations to focus on improving quality, speed, and cost structures. CBA is a useful tool for organizations to evaluate projects in this context as it helps to determine the most cost-effective approach to achieving these goals. By quantifying the costs and benefits in monetary terms, it allows organizations to make informed decisions on which projects to undertake and which to avoid. It also helps organizations to prioritize projects and allocate resources more effectively.

Overall, CBA is a valuable tool for organizations to evaluate the potential benefits and costs of a project and make informed decisions on whether to undertake it or not. It allows organizations to consider the economic viability of the project and how it aligns with their overall goals and objectives.

### F. Balanced Scorecard

The Balanced Scorecard (BSC) is a performance management tool that was developed in the early 1990s to translate an organization's mission and strategy into measurable performance metrics. It was designed to improve strategic measurement and decision making by providing a comprehensive view of an organization's performance. The BSC framework includes four perspectives: financial, customer, internal process, and learning and growth.

One of the great tools that can be used to build key performance indicators (KPIs) for a Balanced Scorecard is SWOT analysis. SWOT analysis is a strategic planning tool that examines an organization's internal and external environment, looking at its strengths, weaknesses, opportunities, and threats. By identifying these factors, organizations can develop KPIs that align with their mission and strategy, and that are focused on improving performance in areas that are most critical to their success.

In summary, the Balanced Scorecard is a performance management tool that helps organizations to translate their mission and strategy into measurable performance metrics, while SWOT analysis is a great tool to build key performance indicators for a Balanced Scorecard. Together, these tools can be used to improve strategic measurement and decision making by providing a comprehensive view of an organization's performance and identifying areas for improvement.

### G. Performance Metrics

Performance metrics are quantitative measurements that are used to evaluate the effectiveness and efficiency of an organization in achieving its goals. They provide a high-level view of the organization's performance and are closely tied to production outputs, as well as business and customer requirements for operational processes.

Performance metrics should be carefully chosen to ensure they are attainable, realistic, and achievable within reasonable timelines. This means that they should be specific, measurable, actionable, relevant and time-bound (SMART). They should also be aligned with the organization's overall goals and objectives, and be able to provide actionable insights that can be used to improve performance.

Overall, performance metrics are an important tool for organizations to evaluate their performance and identify areas for improvement. They are closely tied to production outputs, as well as business and customer requirements for operational processes, and should be chosen carefully to ensure they are attainable, realistic, and achievable within reasonable timelines.

By using performance metrics, organizations can track progress, identify areas for improvement, and make data-driven decisions to optimize their operations and achieve their goals.

### H. Descriptive Analytics

Descriptive analytics is a group of techniques within data mining that enables organizations to gain insights and feedback about their internal performance, operations, and organizational effectiveness by summarizing, generalizing and describing data. These techniques are designed to provide a historical perspective on data, which can be used to identify patterns, trends and anomalies.

Statistics plays a key role in descriptive analytics and several statistical methods can be used to analyze data. Descriptive statistics are used to summarize data and provide a quick overview of the key characteristics of the data. Additionally, sophisticated multivariate data analysis statistical methods such as analysis of variance (ANOVA), K-means clustering, correlation and regression analysis can be used to identify patterns and anomalies in data.

In summary, descriptive analytics is a set of techniques that provide organizations with feedback on their internal performance, operations, and organizational effectiveness. It uses statistics and sophisticated multivariate data analysis statistical methods to summarize, generalize and describe a set of data, identify patterns, and anomalous data, which can help organizations to make data-driven decisions and improve their operations.

### I. Fuzzy logic

Fuzzy logic is a mathematical framework that extends conventional (Boolean) logic to handle the concept of partial truths, which are truth values that fall between completely true and completely false. It allows for reasoning with imprecise or uncertain information, and provides a way to model human decision-making processes. Fuzzy logic is particularly useful in data mining because it allows for the use of qualitative knowledge in the data mining process.

Fuzzy logic is based on the idea that certain concepts or variables can have a range of values that are not limited to the binary true or false. Instead, they can be partially true or false, which allows for a more nuanced representation of reality. This makes fuzzy logic particularly useful in data mining, as it allows for the use of qualitative knowledge such as human intuition, expert judgment, and natural language descriptions in the data mining process.

In summary, Fuzzy logic is a mathematical framework that extends conventional logic to handle the concept of partial truths and reasoning with imprecise or uncertain information. It provides the ability to utilize qualitative knowledge in the data mining process, which can help organizations to make more accurate predictions, identify patterns, and make better decisions based on their data.

### J. Text mining

Text mining is the application of data mining techniques to large textual databases, in order to extract and analyze information and identify patterns. It is a rapidly growing field that enables organizations to gain insights from unstructured data, such as customer feedback, social media posts, and emails.

Text mining uses a variety of techniques to identify keywords and patterns within text fields in data sets. These techniques can include natural language processing, machine learning, and statistical analysis. Text mining applications can also feature textual analysis capabilities that extract and evaluate trends, providing predictive business intelligence that can be used to make informed decisions and plan for the future.

In summary, Text mining is an increasingly popular application of data mining techniques that allows organizations to extract valuable insights from large textual databases. It can identify key words and provide pattern recognition within text fields in data sets, which can help organizations to make more accurate predictions, identify trends, and make better decisions based on their data. The process can also provide predictive business intelligence that can be used to plan for the future.

### K. Predictive Analytics

Predictive analytics is a branch of data mining that uses statistical techniques and models to make predictions about future events or outcomes. It enables organizations to analyze historical data and identify patterns and trends that can be used to forecast future states and make more informed decisions. Predictive analytics can help organizations to better understand and anticipate customer behavior, market trends, and business performance.

The process of predictive analytics involves several steps, including data collection, data cleaning, feature selection, model building, and model evaluation. This process can be applied to various types of data, including structured and unstructured data, and can be used in a wide range of applications, such as customer segmentation, fraud detection, and risk management.

In summary, Predictive analytics is a powerful tool that provides organizations with a view of their future state by forecasting or predicting future events or outcomes. It enables organizations to take appropriate steps in the present to better position themselves for the future by analyzing historical data and identifying patterns and trends that can be used to make more informed decisions. Predictive analytics can help organizations to better anticipate customer behavior, market trends and business performance.

### L. Knowledge Management

Knowledge management is the systematic process of acquiring, organizing, sharing, and leveraging information and knowledge within an organization. It is a holistic approach that involves finding, selecting, distilling, and presenting information in a way that improves the comprehension of a specific subject and facilitates decision-making and action.

The goal of knowledge management is to create and maintain a comprehensive, accurate and up-to-date understanding of an organization's operations, strategies and goals. Knowledge management strategies can include, but not limited to, the use of technology, organizational culture, and processes to identify, create, and share knowledge. The process of knowledge management involves several steps such as knowledge creation, knowledge sharing, and knowledge application.

In summary, Knowledge management is the systematic process of acquiring, organizing, sharing, and leveraging information and knowledge within an organization. It improves the comprehension of a specific subject and facilitates decision-making and action. Knowledge management is a holistic approach that involves finding, selecting, distilling and presenting information in a way that helps organizations to create and maintain a comprehensive, accurate and up-to-date understanding of their operations, strategies and goals.